\documentclass[final,3p,times]{elsarticle}

\usepackage{color}
\usepackage{algorithm}
\usepackage{algorithmic}
\usepackage{tabularray}
\usepackage[normalem]{ulem} 
\usepackage{url}

\usepackage{lineno}
\usepackage{amssymb}
\usepackage{amsmath}

\journal{Nuclear Physics B}

\begin{document}

\begin{frontmatter}

\title{Unsupervised Domain Adaptation with Dynamic Clustering and Contrastive Refinement for Gait Recognition}

\author[SHU]{Xiaolei Liu} 
\author[SHU]{Yan Sun} 
\author[SHU]{Zhiliang Wang}
\author[Soton]{Mark Nixon}

\affiliation[SHU]{organization={School of Computer Engineering and Science, Shanghai University},
	addressline={99 Shangda Road, Baoshan District}, 
	city={Shanghai},
	postcode={200444}, 
	state={Shanghai},
	country={China}}

\affiliation[Soton]{
	organization={School of Electronics and Computer Science, University of Southampton},
	addressline={University Road, Highfield}, 
	city={Southampton},
	postcode={SO17 1BJ}, 
	country={United Kingdom}}

\begin{abstract}
	Gait recognition is an emerging identification technology that distinguishes individuals at long distances by analyzing individual walking patterns. Traditional techniques rely heavily on large-scale labeled datasets, which incurs high costs and significant labeling challenges. Recently, researchers have explored unsupervised gait recognition with clustering-based unsupervised domain adaptation methods and achieved notable success. However, these methods directly use pseudo-label generated by clustering and neglect pseudo-label noise caused by domain differences, which affects the effect of the model training process. To mitigate these issues, we proposed a novel model called GaitDCCR, which aims to reduce the influence of noisy pseudo labels on clustering and model training. Our approach can be divided into two main stages: clustering and training stage. In the clustering stage, we propose Dynamic Cluster Parameters (DCP) and Dynamic Weight Centroids (DWC) to improve the efficiency of clustering and obtain reliable cluster centroids. In the training stage, we employ the classical teacher-student structure and propose Confidence-based Pseudo-label Refinement (CPR) and Contrastive Teacher Module (CTM) to encourage noisy samples to converge towards clusters containing their true identities. Extensive experiments on public gait datasets have demonstrated that our simple and effective method significantly enhances the performance of unsupervised gait recognition, laying the foundation for its application in the real-world. We will release the code at \url{https://github.com/YanSun-github/GaitDCCR} upon acceptance.
\end{abstract}


%

\begin{keyword}
Gait recognition \sep Unsupervised learning \sep Domain adaptation

\end{keyword}

\end{frontmatter}



\section{Introduction}
Gait recognition\cite{yu2006framework} is a biometric technology that identifies individuals by analyzing their walking patterns, such as step length and limb movements. Compared with other biometrics\cite{parashar2023real, azzakhnini2018combining, lin2021xcos, qi2021greyreid, wang2019u} such as fingerprinting, facial recognition, or iris scanning\cite{cao2018automated, sepas2020face, nguyen2017long}, gait recognition has unique advantages: it does not require active participation, can be conducted at a distance, and is difficult to forge. These attributes make gait recognition highly applicable in video surveillance, security systems, and other domains. For gait recognition tasks, a large number of supervised methods\cite{chao2021gaitset, lin2022gaitgl, song2025gait, fan2020gaitpart, rao2023multi, sun2023trigait} have been proposed and achieved good results, but they rely heavily on a large amount of labeled data, which is expensive and difficult to obtain. In addition, this technology is easily affected by clothing changes\cite{singh2018vision, wang2023dygait, chai2021silhouette}, which brings great challenges. 

\begin{figure}[t]
	\centering
	\includegraphics[width=\linewidth]{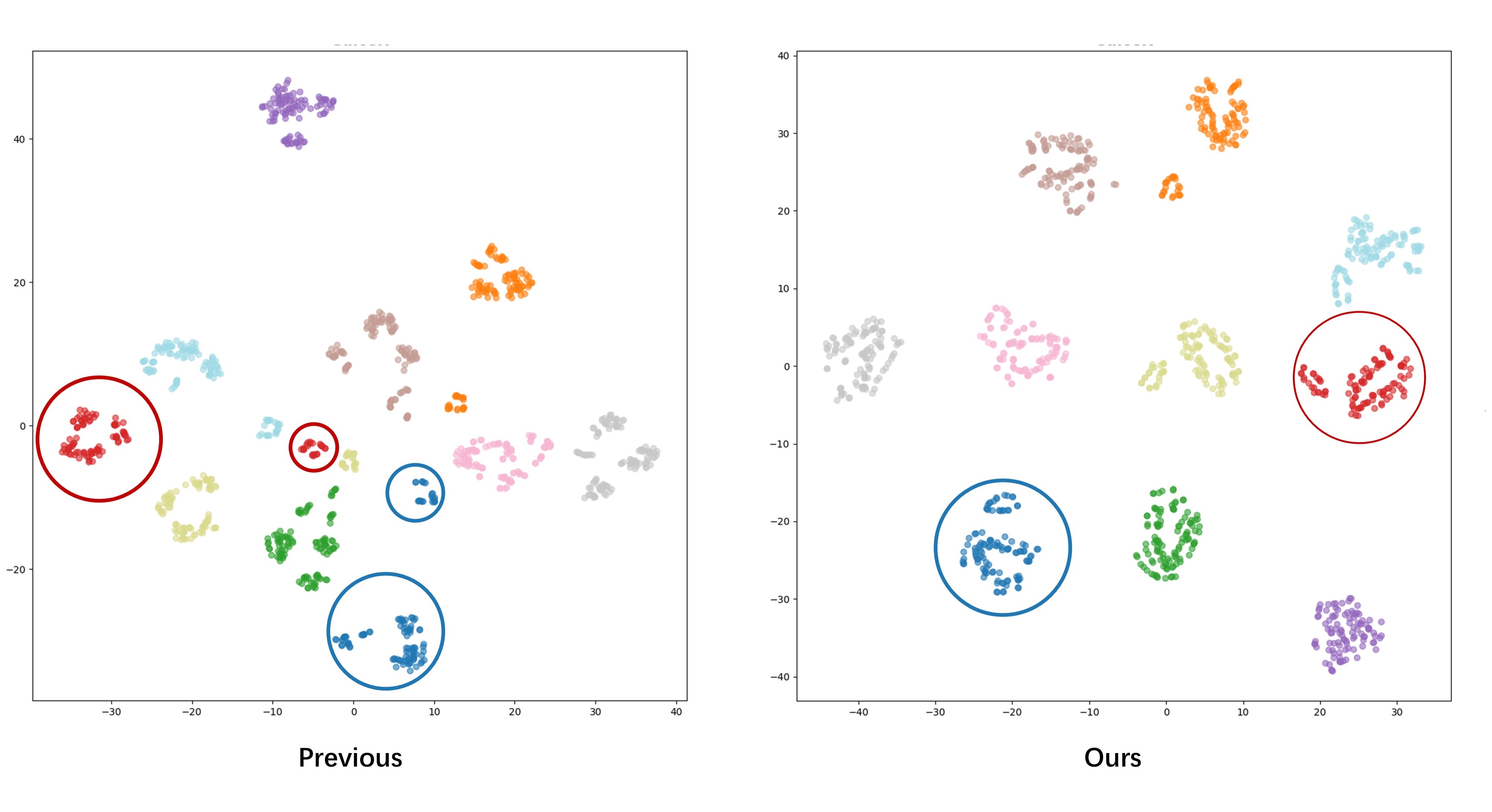}
	\caption{Visualization of clustering results for the previous model and our proposed model. In the figure, samples of different colors represent different subjects, and circles represent clustering results.}
	\label{vis}
\end{figure}
Therefore, in recent years, to facilitate the generalization of this technology, it is necessary to study unsupervised gait recognition techniques. Unsupervised Domain adaptation (UDA)\cite{ganin2015unsupervised} is a technique that transfers labeled source domain knowledge to an unlabeled target domain, which typically consists of two stages: pre-training stage and fine-tuning stage. In the pre-training stage, a general feature extractor is trained on the source domain in a supervised manner. In the fine-tuning stage, the pre-trained encoder is trained on the unlabeled target domain. UDA has been widely applied and has demonstrated significant progress in fields such as pedestrian re-identification\cite{li2018harmonious} and object detection\cite{everingham2010pascal}.

Building on the demonstrated success of UDA in pedestrian re-identification and its notable advantages, several studies\cite{zheng2021trand, habib2023watch, ren2023unsupervised, ma2023fine} have extended its application to gait recognition, primarily focusing on fine-tuning to better adapt to target domains. Among these approaches, clustering-based strategies\cite{ren2023unsupervised, ma2023fine} have shown substantial promise. Generally, these methods comprise two key stages during fine-tuning: clustering and training. The first step involves generating pseudo-labels for the target domain by clustering features extracted from a pre-trained feature encoder. These pseudo-labels are then used for subsequent model training. Recent unsupervised gait recognition methods have reported encouraging results using this clustering-driven domain adaptation strategy. For example, STANet\cite{ma2023fine} enhances feature precision by refining global features through part-based feature extraction, while UGRSF\cite{ren2023unsupervised} employs clothing data augmentation to generate support sets for improved robustness. However, these methods primarily adapt pedestrian re-identification techniques to unsupervised gait recognition without adequately addressing the fundamental differences and unique challenges inherent to the latter. 

Compared to pedestrian re-identification, unsupervised gait recognition presents significantly greater challenges. Firstly, gait is an inherently finer-grained feature, when crossing domains directly, the features extracted by the pre-trained encoder lack discrimination resulting in considerably poorer clustering performance compared to ReID. Secondly, ReID relies on RGB images as input, while gait recognition uses binary silhouette sequences, which are highly sensitive to variations in clothing and viewing angles. This discrepancy aggravates the noise problem of pseudo-labeling, as different samples of the same identity may be incorrectly assigned to different clusters, and a cluster may incorrectly contain multiple identities. Figure \ref{vis} illustrates the clustering results of different subjects in the feature space. The red and blue circled clusters in the left figure demonstrate that previous methods often assigned samples from the same subject to different clusters, leading to noisy pseudo-labels.
\begin{figure}[t]
	\centering
	\includegraphics[width=\linewidth]{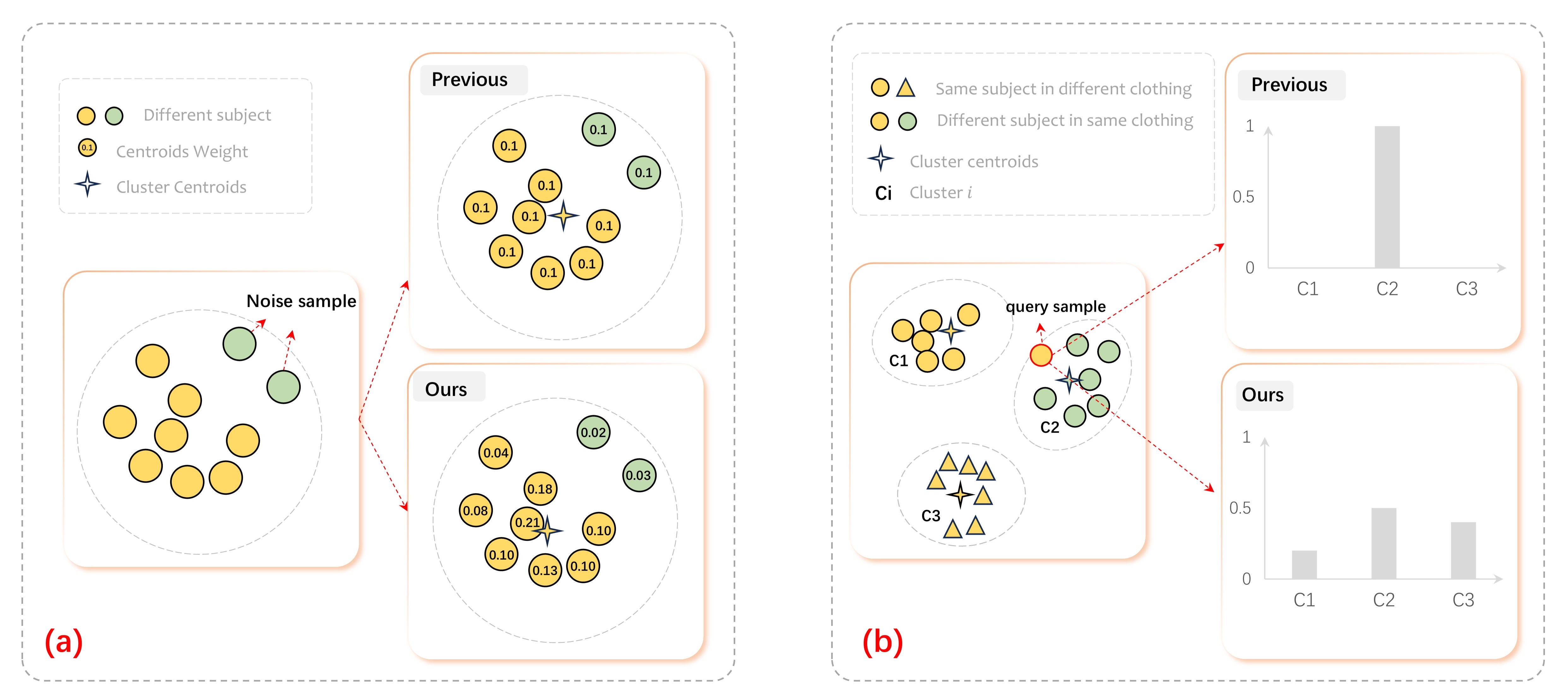}
	\caption{The cluster centroid weighting strategy of previous methods and ours.}
	\label{mo1}
\end{figure}
The main reason for the wrong clustering  is that previous unsupervised gait recognition methods did not adequately address the negative impact of pseudo-label noise, with the direct use of these labels for training, which results in unreliable cluster centroids and suboptimal performance. Clustering-based unsupervised domain adaptation relies heavily on the efficiency of clustering, which in turn relies to some extent on the parameters of the clustering algorithm. The feature distribution is constantly changing during the training process, and the learning ability of the model is gradually improving. However, previous methods frozen the parameters of the clustering algorithm throughout the training process, which may result in a large number of noise samples in the cluster. In addition, when calculating the cluster centroid, it is inevitable that it will be affected by noise, but previous methods calculate centroid by  the average value of the features of all samples in a cluster, as shown in Figure \ref{mo1}(a). Furthermore, the commonly used one-hot label forces all samples to converge towards the assigned cluster centroid, which is detrimental for samples with noisy pseudo-labels, as shown in Figure \ref{mo1}(b).

To address these challenges, we propose an innovative unsupervised gait recognition model called GaitDCCR to enhance clustering accuracy and mitigate the effects of noisy pseudo-labels. The effect of the proposed model is illustrated in the right image of Figure \ref{vis}. It significantly reduces the impact of noisy pseudo-labels, resulting in more reliable clustering outcomes. The blue and red samples, which were originally misclassified into two separate clusters, are accurately grouped together after optimization using our method.

Specifically, in the clustering stage, Dynamic Clustering Parameters(DCP) is proposed to continuously adjust these parameters  to accommodate evolving feature distributions, thereby improving clustering efficiency. Furthermore, assigning a dynamic weight to each sample within a cluster, thereby creating more reliable cluster centroids by summing these weights, as illustrated in Figure \ref{mo1}(a). In the training stage, we employ a classical teacher-student structure\cite{zhang2018deep} to generate soft pseudo-labels. The student network incorporates a Confidence-based Pseudo-label Refinement (CPR) mechanism to assign a confidence score for each sample related to all clusters, as shown in Figure \ref{mo1}(b). This refinement makes that each sample not only converges towards the centroid of its assigned cluster, but is also influenced by other centroids. Meanwhile, in teacher network branch, a Contrast Teacher Module (CTM) is proposed to extract the features of samples with appearance augmentation and find their potential clusters in the feature space, which can address issues that the samples of the same identity is classed into multiple clusters due to variations in clothing or viewing.

Our contributions can be summarized as follows:
\begin{itemize}
	\item{Firstly, in the clustering stage, we innovatively design Dynamic Clustering Parameters (DCP) and Dynamic Weighted Centroids (DWC) to dynamically adjust clustering parameters and calculate cluster centroids by assigning weights.}
	\item{Secondly, in the training stage, the limitations of the one-hot label strategy are addressed by introducing a Confidence-based Pseudo-label Refinement (CPR) method, which encourages noisy sample to align with their true clusters. Additionally, a Contrast Teacher Module (CTM) is implemented to identify the correct clusters for samples with varying clothing conditions.}
	\item{Finally, extensive unsupervised domain adaptation experiments are conducted on two widely used datasets, CASIA-B and OUMVLP. The experimental results unequivocally demonstrate the effectiveness and superiority of our proposed methodology.}
	
\end{itemize}

\section{Related work}
\subsection{Supervised Gait Recognition}
The prevailing supervised gait recognition methods are primarily categorized into three types: model-based, appearance-based, and multi-modal methods. Model-based approaches leverage the physical structure of the body, such as 2D/3D poses and SMPL models. These gait models, typically derived from RGB images, remain unaffected by factors like clothing. For instance, PoseGait\cite{liao2020model} enhanced gait recognition by employing convolutional neural networks to estimate the 3D pose of the human body, thereby extracting spatiotemporal features. GaitGraph\cite{zhang2019graph} integrated skeleton poses with Graph Convolutional Networks (GCN)\cite{zhang2019graph} to establish a contemporary model-based method for gait recognition. GaitSet\cite{chao2021gaitset} innovatively processed the gait sequence as an unordered set, while GaitPart\cite{fan2020gaitpart} focuses on learning features from various body parts to acquire fine-grained information. GaitGL\cite{lin2022gaitgl} proposes a dual-branch structure that considers both global and local features. ParsingGait\cite{zheng2023parsing} introduced a novel gait representation called Gait Parsing Sequence (GPS), which uses fine-grained human segmentation to enhance gait recognition accuracy in real-world scenarios. LandmarkGait\cite{wang2023landmarkgait} proposed an unsupervised landmark discovery network to transform the dense silhouette into a finite set of landmarks with remarkable consistency across various conditions.

Multi-modal gait recognition methods typically utilize multiple gait patterns simultaneously, thus acquiring diverse semantic knowledge. SmplGait\cite{zheng2022gait} learned about 3D views and shapes from the SMPL models' 3D information, while BiFusion\cite{peng2024learning} extracted discriminative gait patterns from the skeleton and integrates them with the contour representation to learn comprehensive gait features. In this study, appearance-based gait recognition models are employed, specifically using  silhouette sequences as input data. 

\subsection{Unsupervised Gait Recognition}
In recent years, there has been considerable advancement in the field of unsupervised domain adaptation for gait recognition. This approach typically involves a two-stage process: pre-training on a labeled source dataset, followed by fine-tuning on an unlabeled target dataset. For instance, TraND\cite{zheng2021trand} employed an entropy-based confidence strategy to minimize distances between similar samples, whereas GOUDA\cite{habib2023watch} introduced an angle-based triplet selection strategy to increase distances between samples with similar viewing angles while reducing distances between samples from different angles. UGRSF\cite{ren2023unsupervised} optimized the memory bank using a multi-cluster updating strategy to address challenges posed by clothing changes. STANet\cite{ma2023fine} refined global features through local features to enhance the model’s generalization capability by combining labeled source domain data with unlabeled target domain data for training. In addition, some researchers proposed domain adaptation for gait recognition, although these are supervised methods.
GaitDAN\cite{huang2024gaitdan} treated the view-change issue as a domain-change issue and proposed an
Adversarial View-change Elimination module equipped with a set of explicit models for recognizing the different gait viewpoints. 
\cite{jaiswal2024domain} introduced a novel domain adaptation technique for practical applications, capitalising on expansive dataset pretraining and precise fine-tuning on targeted, smaller datasets pertaining to specific camera views.

Although these previous methods have achieved commendable results, they fail to effectively address one of the most challenging issues in gait recognition: clothing variations. STANet\cite{ma2023fine} considers that part features are less affected by clothing changes, which are used to refine global features, but the experimental effect is not significant. Similarly, UGRSF\cite{ren2023unsupervised} incorporates a memory bank with clothing-based data augmentation to mitigate the effects of clothing changes but lacks an effective mechanism for refining noisy pseudo-labels. Experimental results also show that these methods have certain effects when clothing changes, but the improvement is not significant. Furthermore, these approaches rely directly on clustering-generated pseudo-labels for training, overlooking the pervasive issue of noisy pseudo-labels. Distinct from these prior works, our proposed method combines data augmentation and a teacher-student structure to address the challenges posed by clothing variations while simultaneously refining hard pseudo-labels. This dual strategy significantly enhances the performance of unsupervised gait recognition and demonstrates the necessity of effectively resolving pseudo-label noise and clothing variation to achieve superior results.

\subsection{Unsupervised Domain Adaptation}
Unsupervised domain adaptation (UDA) is a crucial machine learning technique that enables the adaptation of a model from a source domain to a target domain without requiring labeled data from the target domain. This technique is particularly beneficial when addressing issues such as data privacy, security, or storage limitations, as it enables model migration and optimization without the need for source domain data. Collaborative class conditional generative adversarial networks (CCGAN)\cite{li2020model} enhanced prediction models by generating data in the target style without relying on source data. DANCE\cite{saito2020universal} achieved domain adaptation through self-supervised learning, utilizing neighborhood clustering techniques to learn the target domain's structure and entropy-based feature alignment to manage unknown categories. Multi-source UDA\cite{ren2022multi} added complexity by addressing domain offsets both between and within source domains. GSDE\cite{westfechtel2024gradual} improved model accuracy in the target domain by gradually enlarging the source domain dataset and employing pseudo-source samples to enhance domain alignment during early training stages.
\subsection{Contrastive Learning}
Contrastive learning, a significant branch of self-supervised learning, has enjoyed considerable interest in recent years. This approach examines the intrinsic structure and characteristics of data by learning data representation without the use of labeled data. SimCLR\cite{chen2020simple}, for instance, learned visual representations by maximizing the similarity of positive pairs and minimizing the similarity of negative pairs. This highlights the importance of data augmentation techniques. However, MoCo\cite{he2020momentum} introduced a dynamic memory bank that allows access to a large number of negative samples at each training step, thereby improving the learning efficiency and the quality of the representation. In contrast, BYOL\cite{grill2020bootstrap} demonstrated a contrastive learning strategy that does not require negative examples. Specifically, two networks learn a representation by predicting each other's output, thereby demonstrating that it is possible to learn effectively even without direct negative comparison. SwAV\cite{caron2020unsupervised} employed a method of exchanging cluster assignments of different views of the same image, thereby reducing the reliance on a large number of negative samples. Given the excellent performance of contrastive learning in self-supervised learning, contrastive learning is also designed in our model to address the issue of objects being clustered into different classes due to clothing changes.

\section{Baseline}
Let $X_s = \left \{ x_{1}, x_{2}, \dots x_{N_{s}} \right \}, X_t = \left \{ x_{1}, x_{2}, \dots x_{N_{t}} \right \}$ represent a labeled source domain dataset and an unlabeled target domain dataset, respectively, where  $x_{i}$ represents the  $i^{th}$ sample,  $N_{s}$ is the number of samples in the source domain, and  $N_{t}$  is the number of samples in the target domain. The purpose of unsupervised domain adaptation for gait recognition is to train a feature extractor that transfers knowledge from the source domain to the target domain, i.e. training the feature extractor on the labeled source domain and then using unsupervised methods to train it on the unlabeled target domain. Inspired by cluster-contrast\cite{dai2022cluster}, we use cluster-based contrastive learning for unsupervised training, which consists of two main stages:

\textbf{Stage 1: Clustering.} At the beginning of each epoch, the target domain sample features are mapped to the feature space using the pre-trained feature extractor $f$, then the clustering algorithms assign pseudo labels to each sample, obtaining  $\left \{ (x_{1},y_{1}), (x_{2},y_{2}), \dots (x_{N_{t}, }y_{N_{t}}) \right \}, y_{i} \in \left\{ 1, \dots, C\right\}$, where $y_{i} \in \left\{ 1, \dots, C\right\}$ represents the pseudo label for the $i^{th}$ sample, and $C$ is the total number of clusters. Concurrently, a memory bank is initialized by the average of all samples in each cluster as the preliminary cluster centroids, which are formulated as follows:
\begin{equation}
	\label{3.1}
	m_{i} = \frac{1}{\left | C^{i} \right | }\sum_{i  \in C^{i}}^{}f_{i} 
\end{equation}
where  $f_{i}$ represents the feature of the  $i^{th}$ sample,  $C^{i}$ indicates the cluster of the $i^{th}$ sample and $\left | C^{i} \right |$ denotes the number of samples in the cluster. 

\textbf{Stage 2: Training.} The backbone is updated using the InfoNCE loss, defined as follows:
\begin{equation}
	\label{3.2}
	\mathcal L_{InfoNCE} = -\sum_{i=1}^{N}log\frac{exp(f_{i} \cdot m_{+}/\tau )}{\sum_{j=1}^{C} exp(f_{i} \cdot m_{j}/\tau)}
\end{equation}
where $m_{+}$ represents the centroid of the cluster to which the $i^{th}$ sample belongs, $m_{j}$ represents the $j$-th centroid in the memory bank, and $\tau$ is a temperature parameter. The dot product $(\cdot)$ indicates cosine similarity. Moreover,  a momentum update strategy is used to update the cluster centroids:

\begin{equation}
	\label{3.3}
	m_{i} =\mu \cdot m_{i} + (1-\mu)\cdot f_{i}
\end{equation}
where $\mu \in [0,1]$ is the momentum factor.

\section{Method}
\subsection{Overview}
\begin{figure}[t]
	\centering
	\includegraphics[width=\linewidth]{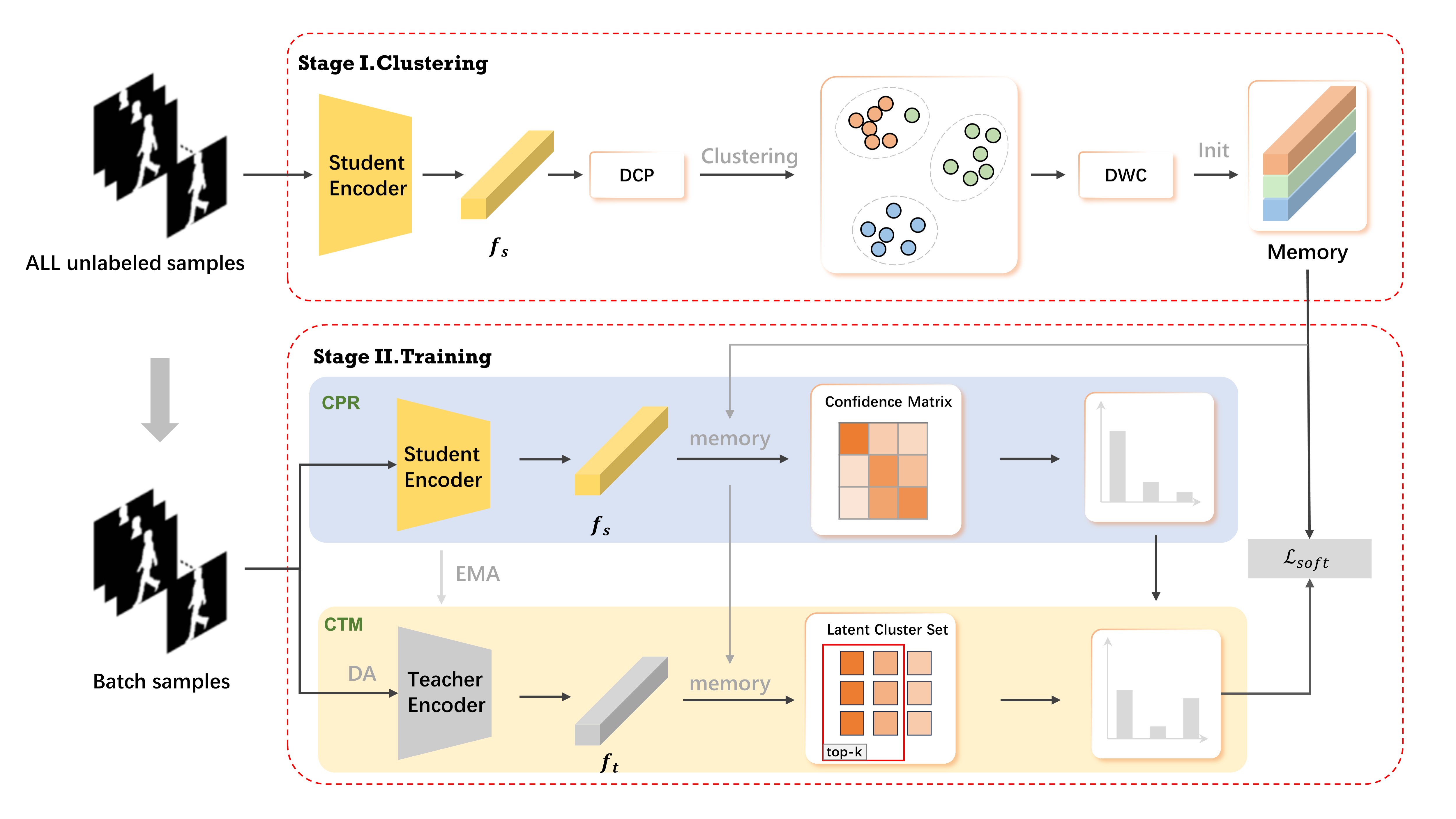}
	\caption{The framework of our model. “DA” stands for Data Augmentation, and “EMA” represents Exponential Moving Average, which is the parameter update method of the teacher encoder.}
	\label{model}
\end{figure}

In this paper, we propose a model named GaitDCCR for unsupervised gait recognition. The framework
of our method is in Figure \ref{model}, and the pseudo-code is shown in Algorithm \ref{alg1}. In the clustering stage, Dynamic Clustering Parameters (DCP) is proposed to dynamically change clustering parameters to improve clustering accuracy. Moreover, we introduce Dynamic Weight Centroid (DWC) to recalculate the cluster centroids by assigning weights to each sample and obtain reliable clustering centroids. In the training phase, we use the classical Teacher-student structure. In the student branch, we use Confidence-based Pseudo-label Refinement (CPR) to adaptively move noisy samples to their true clustering centroids based on the confidence matrix. In the teacher branch, a Contrastive Teacher Module (CTM) is designed by incorporating contrastive learning to identify the potential true clusters of samples from the Latent Cluster Set. The details of our model are described as follows.

\subsection{Dynamic Clustering Parameters(DCP)}
For clustering-based unsupervised domain adaptation methods, the quality of clustering results is a critical determinant of overall performance. In clustering algorithms, the probability threshold $\epsilon$ plays a pivotal role in determining clustering results. If the $\epsilon$ is too small, more noise samples will be included in the cluster, if the $\epsilon$ is too large, more real samples will be excluded. As the performance of the model gradually improves with training, the ability to extract discriminative features is poor in the early stage of training. 

Therefore, unlike the previous method that uses a fixed $\epsilon$ throughout the training process, we propose a dynamic clustering parameter(DCP) method to dynamically adjust the $\epsilon$. In the early stage of training, a larger $\epsilon$ is used to reduce the impact of noise samples. As the model's discriminative ability improves, the $\epsilon$ gradually decays, so that the proportion of real samples in the cluster gradually increases. Specifically, we adopt an exponentially decaying strategy, formalized as:
\begin{equation}
	\label{dcp}
	\epsilon=\epsilon_{0} \cdot \eta^{epochs}
\end{equation}
where $\epsilon_{0}$ is the initial $\epsilon$ value and $\eta$ represents the decay rate, and $epochs$ stands for the value of epochs during training. 

\subsection{Dynamic Weighted Centroids (DWC)}
In clustering-based unsupervised domain adaptation methods, cluster centroids are typically computed and stored in a memory bank. However, due to the inevitable presence of noisy samples in the clusters, these centroids may lack accuracy, which may affect the performance of the model. Most previous methods regard the average features of all samples  in the cluster as the centroid, which is greatly affected by noise samples. To address this issue, we propose a novel approach for computing cluster centroids by incorporating dynamic weighting based on the distribution of samples within each cluster.

Empirical analysis shows that noisy samples tend to be distributed at the edge of clusters, while correctly clustered samples are more likely to be concentrated near the cluster center. Based on this observation, we define a metric called density distance, which measures the average distance of a sample to all other samples within the same cluster:

\begin{equation}
	\label{4.1.1}
	D_{i} = \sigma(-\frac{1}{\left | C^{i} \right |}\sum_{j\in C^{i}}d(i,j)) 
\end{equation}
$d(i,j)$ stands for the distance between the $i^{th}$ sample and the $j^{th}$ sample, and  $\sigma$ is the activation function. According to the density distance, we can distinguish between noise samples and real samples, that is, samples with smaller density distance are more likely to be real samples, while samples with larger density distance are likely to be noise samples.

To mitigate the influence of noisy samples, we use the density distance to assign dynamic weights to samples when calculating cluster centroids. The weight $w_{i}$ for a given sample is computed as follows:

\begin{equation}
	\label{4.1.2}
	w_{i}=\frac{D_{i}}{\sum_{i \in C^{i}}{D_{i}}} 
\end{equation}
Using these weights, the new cluster centroids are calculated as a weighted sum of sample features:
\begin{equation}
	\label{4.1.3}
	m_{i}' = \sum_{x_{i} \in C_{i}}w_{i}f_{i}
\end{equation}
where $m_{i}'$ represents the newly cluster centroid, and $w_{i}$ is the confidence score of these samples. By assigning greater weights to samples closer to the cluster center and smaller weights to those further away, our method produces more reliable cluster centroids. When this optimization process is combined with Dynamic Clustering Parameters (DCP), the robustness and accuracy of the clustering stage can be significantly enhanced, especially during early training when model capacity is still limited.

\subsection{Confidence-Based Pseudo Label Refinement (CPR)}
To further mitigate the detrimental effects of noisy pseudo-labels, we propose a Confidence-Based Pseudo-Label Refinement (CPR) mechanism inspired by prior research\cite{miao2024confidence}. Traditional methods typically use hard pseudo-labels generated by clustering to directly train the model, forcing samples to align with their assigned cluster centroids while ignoring the possibility that other nearby centroids may better represent the true clusters. In order to make the sample flexible and close to other close clusters, we propose CPR to refine hard pseudo-labels into soft pseudo-labels by assigning a confidence score to each sample for all clusters based on the distance, so that the sample is close to the assigned cluster and also close to other clusters with similar distances. Specifically, a confidence matrix denoted by $F$ is employed to store the confidence of each sample for all clusters. This matrix has dimensions $R^{B\times C}$, where $B$ and $C$ represent the number of samples and clusters, respectively. The confidence associated with each sample is defined as:
\begin{equation}
	\label{4.2.1}
	F_{i,j}=\frac{p_{i,j}}{\sum_{j=1}^{C}p_{i,j} }, p_{i,j} = \sigma (-d(i,j)) 
\end{equation}
where $d(i,j)$ denotes the distance between the $i^{th}$ sample and the $j^{th}$ cluster, $\sigma$ is the activation function, and $C$ is the total number of clusters.  The refined pseudo-labels $\hat{y_{i}}$ are then computed by combining the original one-hot pseudo-labels with the confidence matrix.
\begin{equation}
	\label{4.2.2}
	\hat{y_{i}} = \alpha \cdot y_{i} + (1- \alpha)\cdot F_{i} 
\end{equation}
where $F_{i} \in R^{C}$ represents the confidence that the $i^{th}$ sample belongs to all clusters, $y_{i}$ represents the one-hot pseudo label, $\alpha$ is a hyperparameter and $\alpha \in \left[0,1 \right]$ . This refinement enables real samples to align more closely with their assigned cluster centroids while guiding noisy samples toward their true clusters.

By dynamically incorporating sample-level confidence, CPR substantially enhances the reliability of pseudo-labels, reducing the negative impact of noise and improving the overall training efficacy. This mechanism plays a pivotal role in the training phase, ensuring that the model effectively learns distinguishing feature.

\begin{figure}[t]
	\centering
	\includegraphics[width=\linewidth]{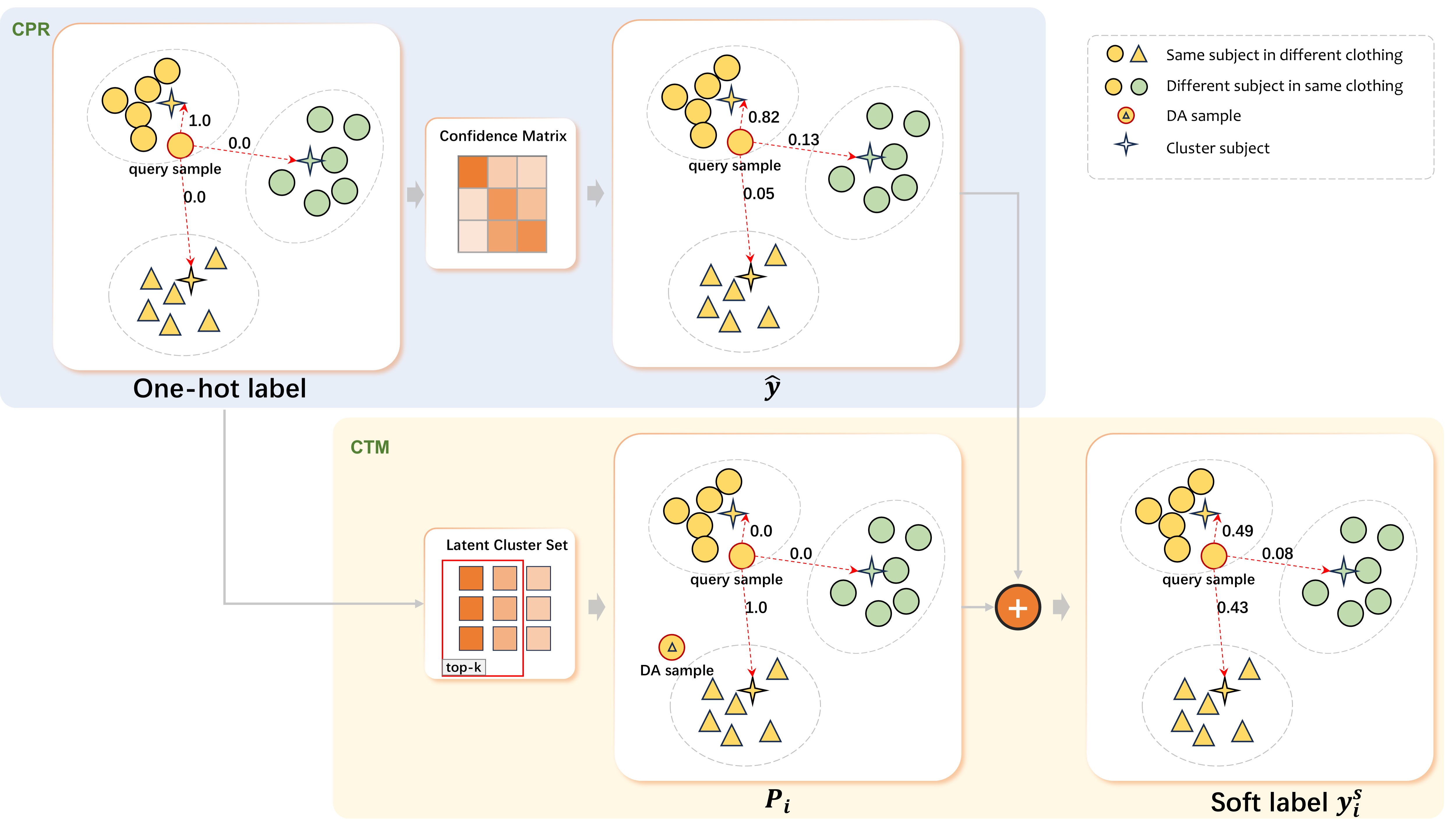}
	\caption{The generation process of soft pseudo-labels. “DA sample” is obtained by query sample through data augmentation, which simulates the change of clothing.}
	\label{CPR}
\end{figure}
\subsection{Contrastive Teacher Module(CTM)}
Given that gait recognition is a fine-grained task sensitive to clothing variations, we have identified that this sensitivity can lead to samples from the same identity being assigned to multiple clusters due to clothing changes. This issue has been largely overlooked in previous works. To address this challenge, we introduce a novel approach combining contrastive learning with newly proposed data augmentation techniques to refine pseudo-labels, thereby narrowing the distance between samples of the same identity with different clothing conditions. Additionally, we adopt a classical teacher-student network structure, which enables the model to incorporate prior knowledge and smooths the learning process, as depicted in Figure \ref{CPR}.

\subsubsection{Data augmentation}
Inspired by previous data augmentation methods\cite{fan2023learning} in gait recognition, we propose a new data augmentation method to augment the input batch with binary gait silhouette data to simulate various clothing conditions during the training phase. Specifically, our data augmentation technique divides the human body into three regions—head, body, and legs—at the proportions of 1/4, 3/4, and 1/2, respectively. The body region is then subject to dilation and erosion operations. Previous data augmentation methods have manipulated all three parts of the body with certain probabilities, but in reality, the head and legs are minimally affected by clothing changes. Thus, our approach focuses solely on manipulating the body region, which is more sensitive to clothing variations.

\begin{figure}[t]
	\centering
	\includegraphics[width=\linewidth]{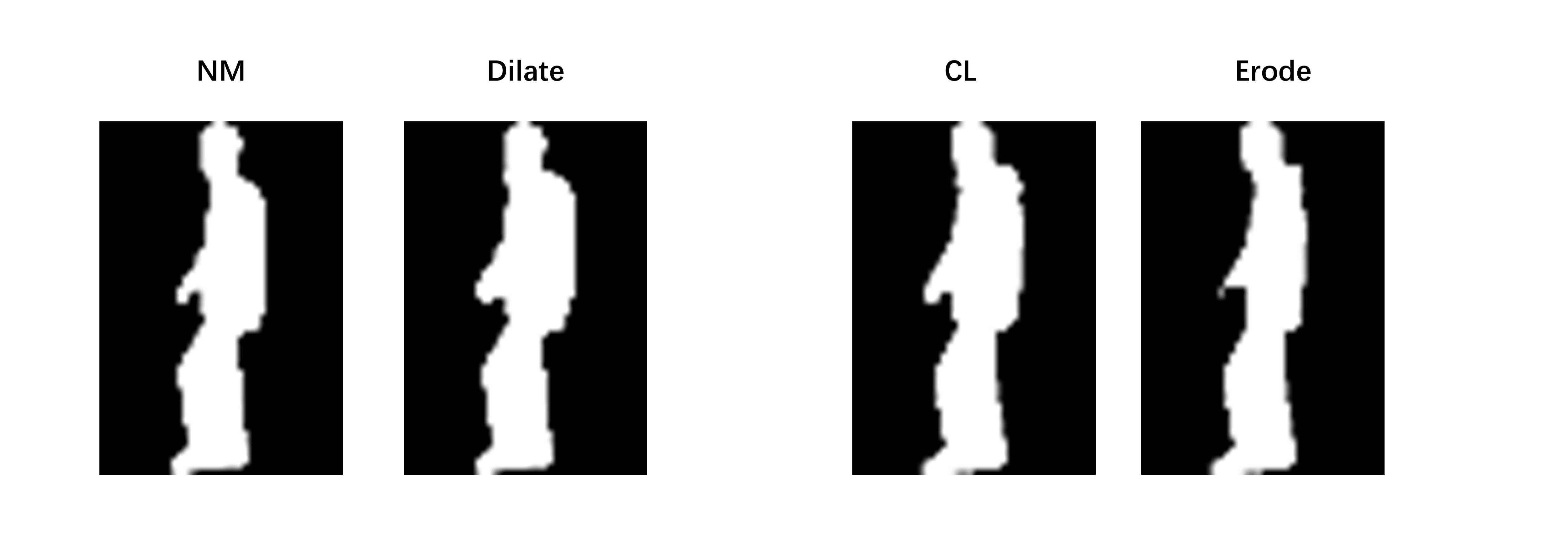}
	\caption{The visualization of data augmentation on NM and CL conditions.}
	\label{DA}
\end{figure}

Figure \ref{DA} illustrates the effects of our data augmentation method. For instance, dilation operations applied to the "NM" condition can simulate clothing changes similar to those seen in the "CL" condition, while erosion applied to the "CL" condition mimics clothing changes that would be observed in the "NM" condition. This approach captures realistic clothing changes under simpler conditions and serves as a proof of concept. More complex and varied clothing scenarios will be considered to further improve the robustness of our model.

\subsubsection{Smoothed soft pseudo-labels}
To address the effect of clothing variation on model training, we combine the data enhancement methods described above with contrast learning to compute a smooth soft pseudo-label for each sample. Specifically, the unlabeled gait data $x_{i}$ is first processed using the data augmentation approach described above to generate samples $\tilde{x}_i$ that simulate different dressing situations, which are fed into the teacher network to extract features and obtain  $f_{t}$. Next, we look for the k closest clusters to  $f_{t}$ in the feature space and create a Latent Cluster Set to be stored. These $k$ clusters are considered to be potential clusters that have the same identity as  $x_{i}$ , and our goal is to bring  $x_{i}$ closer to these $k$ clusters.

To optimize the robustness of labels to appearance changes, we minimize the distance between $x_{i}$ and these $k$ clusters. Specifically, We assign probabilities to the $k$ closest clusters to $x_{i}$ by normalization, while the probability score for the remaining clusters is set to 0. This approach can be formulated as follows:
\begin{equation}
	\label{4.4.1}
	P_{i}=[y_{i}^{0},\dots , y_{i}^{j}, \dots, y_{i}^{C}], y_{i}^{j} = \begin{cases}
		\frac{d(\tilde{x_{i}}, C_{j})}{\sum f_{C_{i}^{k}}} , & \text{if } j \in C_{i}^{k} \\
		0, & \text{otherwise}
	\end{cases}
\end{equation}
where $y_{i}^{j}$ indicates the probability score of sample $x_{i}$  belonging to the $j$-th cluster, and $C_{i}^{k}$ represents the k-nearest clusters to sample $x_{i}$. The final soft pseudo-label can be obtained by combining $P_{i}$ and $\hat{y_{i}}$:
\begin{equation}
	\label{4.4.2}
	y_{i}^{s} = \beta \cdot P_{i} + (1- \beta)\cdot \hat{y_{i}}
\end{equation}

It's important to note that the teacher encoder is not trained directly but is updated through an exponential moving average:
\begin{equation}
	\theta_{t}^{T}=\gamma \theta_{t}^{T-1} + (1-\gamma)\theta_{s}^{T}, \theta_{t}^{0}=\theta_{s}^{0}
\end{equation}
where $\theta _{s}$ represents the parameters of the student network, $\theta _{t}$ represents the parameters of the teacher network, $T$ denotes iteration number, and $\gamma \in [0,1]$.

Figure \ref{CPR} shows the process of optimizing the one-hot label to the soft label. In the student branch, the confidence matrix is obtained by calculating the similarity with all clusters, and then the confidence-based pseudo label $y_{i}^{s}$ is calculated according to Equation \ref{4.2.2}. In the teacher branch, the query sample is  augmented to get DA sample, and the Latent Cluster Set stores the k clusters closest to the DA sample. $P_{i}$ is obtained according to Equation \ref{4.4.2}, $k=1$ in Figure \ref{CPR} . Finally, the soft label $y_{i}^{s}$ is obtained by combining $y_{i}^{s}$ and $P_{i}$.
\subsection{Loss function}
To fully optimize the performance, we compute the InfoNCE loss using the obtained soft pseudo-labels and the memory bank. Specifically, the total loss function of the model is defined by the following equation:
\begin{equation}
	\label{4.3.1}
	\mathcal L_{soft} = \frac{1}{N} \sum_{i=1}^{N}[l_{ce}(M_{q}f_{i}, y_{i}^{s} )] 
\end{equation}
where $ l_{ce}$ denotes the cross-entropy loss and $M_{q}$ denotes the clustering features stored in the memory bank. 
\begin{algorithm}[t]
	\begin{minipage}{.95\linewidth}
		\renewcommand{\algorithmicrequire}{\textbf{Input:}}
		\renewcommand{\algorithmicensure}{\textbf{Output:}}
		\caption{The training procedure of GaitDCCR}
		\label{alg1}
		\begin{algorithmic}[1]
			\REQUIRE Unlabeled target dataset $X_{N_{t}}$, pretrained feature extractor $f$, epochs, iterations, hyperparameters
			\ENSURE Feature extractor $f$
			\FOR{epoch in range(epochs)}
			\STATE \textbf{Clustering stage:}
			\STATE Extracting $f_{s}$ for all samples by $f$;
			\STATE Dynamic clustering parameters: $\epsilon = \epsilon_{0} \cdot \eta^{epoch}$;
			\STATE Getting one-hot labels $y_{i}$ by clustering;
			\STATE Computing the clustering centroid $m'$ by DWC;
			\STATE Initialize memory bank with $m'$;
			\STATE
			\STATE \textbf{Training stage:}
			\FOR{iter in range(iterations)}
			\STATE Extracting $f_{s}$ for batch by student encoder;
			\STATE Computing the confidence matrix $F$;
			\STATE Computing the confidence-based pseudo-label:$y_{i}^{c} = \alpha \cdot y_{i} + (1- \alpha) \cdot F_{i}$;
			\STATE Extracting  $f_{t}$ for batch by teacher encoder;
			\STATE Computing Latent Cluster Set 
			\STATE Getting pseudo-label $P_{i}$;
			\STATE Computing soft pseudo-label: $y_{i}^{s} = \beta \cdot y_{i}^{k} + (1- \beta) \cdot \hat{y_{i}}$;
			\STATE Computing InfoNCE loss;
			\STATE Updating student encoder, teacher encoder;
			\STATE Updating memory bank;
			\ENDFOR
			\ENDFOR
		\end{algorithmic} 
	\end{minipage}
\end{algorithm}
\section{Experiments}
This study conducts a series of experiments to verify the effectiveness of proposed unsupervised gait recognition model. We conducted domain adaption experiments on three widely used datasets, including OUMVLP to CASIA-B, CASIA-B to OUMVLP, and GREW to OUMVLP.

\subsection{Dataset and Implementation Details}
\textbf{CASIA-B\cite{yu2006framework}:} This dataset is one of the most popular in gait recognition. It includes 124 subjects under three walking conditions: normal walking (NM), walking with a bag (BG), and walking in a coat (CL). The dataset is captured by 11 cameras positioned from 0° to 180°, offering a diverse range of viewpoints. Each subject has a total of 110 sequences, with the first 74 subjects used for training and the remaining 50 for testing. During testing, the first four "NM" conditions are used as the gallery, while the remaining conditions serve as probes.

\textbf{OUMVLP\cite{takemura2018multi}:} This large indoor gait dataset covers a wide range of genders and ages with 10,307 subjects, providing 14 different viewing angles (0°-90°, 180°-270°), each containing two sequence sets (01 and 02). In our experiments, 5,153 subjects are used for training and 5,154 subjects for testing. During testing, the "01" sequences serve as probes, and the "02" sequences as the gallery.

\textbf{GREW\cite{zhu2021gait}:} A large-scale, in-the-wild dataset comprising 26,345 subjects and 128,671 sequences, including four data types: Silhouettes, Optical Flow, 2D Pose, and 3D pose. The dataset is divided into three parts: training (20,000 identities and 102,887 sequences), validation (345 identities and 1,784 sequences), and testing (6,000 identities with four sequences each). In our experiments, we use this dataset as the source dataset for the pre-trained feature extractor.

\textbf{Implementation Details}: We utilize two different gait backbones (GaitSet and GaitGL) as our gait feature extractors, following the implementation from the OpenGait\cite{fan2023opengait} framework. All datasets are adjusted to a uniform silhouette size of 64x44, and we select 30 frames for training, with all frames used for evaluation. The Adam optimizer is employed with a learning rate of 1e-4 and a weight decay of 5e-4. The momentum $\mu$ is set at 0.2 and $\gamma$  is set at 0.99. For the  hyperparameters, $\epsilon_{0}=0.8$, $\eta=0.97$, $\alpha=0.4$, $\beta$=0.4. Other parameters are shown in Table \ref{setting}.

\begin{table}[t]
\centering
\caption{Settings for our experiments on the CAISA-B and OUMVLP datasets}
\label{setting}
\begin{tblr}{
  width = \linewidth,
  colspec = {Q[127]Q[142]Q[229]Q[229]Q[204]},
  cell{1}{1} = {r=2}{},
  cell{1}{2} = {r=2}{},
  cell{1}{3} = {c=3}{0.662\linewidth,c},
  cell{3}{1} = {r=4}{},
  cell{7}{1} = {r=4}{},
  hline{1,3,7,11} = {-}{},
  hline{2} = {3-5}{},
}
Backbone & Setting      & Domain Adaption &                &             \\
         &            & OU $\rightarrow$ CA  & CA $\rightarrow$ OU & GR $\rightarrow$ OU \\
GaitSet  & Batchsize  & (16,8)          & (16,8)         & (16,8)      \\
         & Epoch      & 25              & 25             & 25          \\
         & Iteration  & 200             & 300            & 400         \\
         & MileStones & {[}2K, 4K]      & {[}3K, 6K]     & {[}4K, 8K]  \\
GaitGL   & Batchsize  & (16,8)          & (16,8)         & (16,8)      \\
         & Epoch      & 15              & 25             & 25          \\
         & Iteration  & 200             & 300            & 400         \\
         & MileStones & {[}1K, 2K]      & {[}3K, 6K]     & {[}4K, 8K]  
\end{tblr}
\end{table}

\definecolor{Alto}{rgb}{0.87,0.87,0.87}
\begin{table}
	\centering
	\caption{Rank-1 (\%) comparison on CASIA-B. The best results of the model with different backbones and clothing conditions are highlighted in bold, and the second are marked with underline.}
	\fontsize{8pt}{9pt}\selectfont
	\label{CAISA-B}
	\begin{tblr}{
			width = \linewidth,
			colspec = {Q[88]Q[92]Q[160]Q[54]Q[54]Q[54]Q[54]Q[54]Q[54]Q[56]Q[56]Q[56]Q[56]Q[56]Q[63]},
			row{3} = {Alto,c},
			row{16} = {Alto,c},
			cell{1}{1} = {r=2}{},
			cell{1}{2} = {r=2}{},
			cell{1}{3} = {r=2}{},
			cell{1}{4} = {c=11}{0.604\linewidth,c},
			cell{1}{15} = {r=2}{},
			cell{3}{1} = {c=15}{0.926\linewidth},
			cell{4}{1} = {r=4}{},
			cell{4}{2} = {r=4}{},
			cell{8}{1} = {r=4}{},
			cell{8}{2} = {r=4}{},
			cell{12}{1} = {r=4}{},
			cell{12}{2} = {r=4}{},
			cell{16}{1} = {c=15}{0.926\linewidth},
			cell{17}{1} = {r=12}{},
			cell{17}{2} = {r=4}{},
			cell{21}{2} = {r=4}{},
			cell{25}{2} = {r=4}{},
			cell{29}{1} = {r=12}{},
			cell{29}{2} = {r=4}{},
			cell{33}{2} = {r=4}{},
			cell{37}{2} = {r=4}{},
			hline{1,3-4,8,12,16-17,29} = {-}{},
			hline{21,25,33,37,41} = {2-15}{},
		}
		Backbone                                                                             & Condition & Method  & Probe View &      &      &      &      &      &      &      &      &      &      & Mean          \\
		&           &         & 0°         & 18°  & 36°  & 54°  & 72°  & 90°  & 108° & 126° & 134° & 162° & 180° &               \\
		Fully supervised methods with all labels on CASIA-B                                  &           &         &            &      &      &      &      &      &      &      &      &      &      &               \\
		& NM        & GaitNet\cite{zhang2019gait} & 91.2       & 92.0 & 90.5 & 95.6 & 86.9 & 92.6 & 93.5 & 96.0 & 90.9 & 88.8 & 89.0 & 91.6          \\
		&           & GaitSet\cite{chao2021gaitset} & 90.8       & 97.9 & 99.4 & 96.9 & 93.6 & 91.7 & 95.0 & 97.8 & 98.9 & 96.8 & 85.8 & 95.0          \\
		&           & MT3D\cite{lin2020gait}    & 95.7       & 98.2 & 99.0 & 97.5 & 95.1 & 93.9 & 9.1  & 98.6 & 99.2 & 98.2 & 92.0 & 96.7          \\
		&           & GaitGL\cite{lin2022gaitgl}  & 96.0       & 98.3 & 99.0 & 97.9 & 96.9 & 95.4 & 97.0 & 98.9 & 99.3 & 98.8 & 94.0 & 97.4          \\
		& BG        & GaitNet\cite{zhang2019gait} & 83.0       & 87.8 & 88.3 & 93.3 & 82.6 & 74.8 & 89.5 & 91.0 & 86.1 & 81.2 & 85.6 & 85.7          \\
		&           & GaitSet\cite{chao2021gaitset} & 83.8       & 91.2 & 91.8 & 88.8 & 83.3 & 81.0 & 84.1 & 90.0 & 92.2 & 94.4 & 79.0 & 87.2          \\
		&           & MT3D\cite{lin2020gait}    & 91.0       & 95.4 & 97.5 & 94.2 & 92.3 & 86.9 & 91.2 & 95.6 & 97.3 & 96.4 & 86.6 & 93.0          \\
		&           & GaitGL\cite{lin2022gaitgl}  & 92.6       & 96.6 & 96.8 & 95.5 & 93.5 & 89.3 & 92.2 & 96.5 & 98.2 & 96.9 & 91.5 & 94.5          \\
		& CL        & GaitNet\cite{zhang2019gait} & 42.1       & 58.2 & 65.1 & 70.7 & 68.0 & 70.6 & 65.3 & 69.4 & 51.5 & 50.1 & 36.6 & 58.9          \\
		&           & GaitSet\cite{chao2021gaitset} & 61.4       & 75.4 & 80.7 & 77.3 & 72.1 & 70.1 & 71.5 & 73.5 & 73.5 & 68.4 & 50.0 & 70.4          \\
		&           & MT3D\cite{lin2020gait}    & 76.0       & 87.6 & 89.8 & 85.0 & 81.2 & 75.7 & 81.0 & 84.5 & 85.4 & 82.2 & 68.1 & 81.5          \\
		&           & GaitGL\cite{lin2022gaitgl}  & 76.6       & 90.0 & 90.3 & 87.1 & 84.5 & 79.0 & 84.1 & 87.0 & 87.3 & 84.4 & 69.5 & 83.6          \\
		Unsupervised Domain Adaptation methods without any labels on CASIA-B (OUMVLP  $\rightarrow$ CASIA-B) &           &         &            &      &      &      &      &      &      &      &      &      &      &               \\
		GaitSet                                                                              & NM        & GOUDA\cite{habib2023watch}   & -          & -    & -    & -    & -    & -    & -    & -    & -    & -    & -    & 87.1          \\
		&           & UGRSF\cite{ren2023unsupervised}   & 85.2       & 93.6 & 96.4 & 93.8 & 90.0 & 84.6 & 89.6 & 92.3 & 96.9 & 93.2 & 77.4 & 90.3          \\
		&           & STAGait\cite{ma2023fine} & 84.7       & 95.8 & 97.2 & 95.8 & 95.1 & 92.9 & 94.8 & 96.1 & 97.3 & 96.6 & 79.2 & \textbf{93.2} \\
		&           & Ours    & 76.7       & 93.8 & 95.4 & 94.1 & 89.5 & 85.2 & 87.6 & 91.1 & 93.1 & 89.5 & 78.1 & \uline{90.4}  \\
		& BG        & GOUDA\cite{habib2023watch}   & -          & -    & -    & -    & -    & -    & -    & -    & -    & -    & -    & 68.1          \\
		&           & UGRSF\cite{ren2023unsupervised}   & 78.5       & 88.3 & 89.8 & 88.0 & 83.5 & 76.4 & 80.5 & 83.5 & 85.8 & 84.8 & 72.9 & \uline{82.9}  \\
		&           & STAGait\cite{ma2023fine} & 77.5       & 87.5 & 90.0 & 89.2 & 86.2 & 81.2 & 87.1 & 89.9 & 94.1 & 91.1 & 70.9 & \textbf{85.9} \\
		&           & Ours    & 73.5       & 80.2 & 83.3 & 80.8 & 75.8 & 69.4 & 72.5 & 81.1 & 79.3 & 79.6 & 62.3 & 80.1          \\
		& CL        & GOUDA\cite{habib2023watch}   & -          & -    & -    & -    & -    & -    & -    & -    & -    & -    & -    & 27.2          \\
		&           & UGRSF\cite{ren2023unsupervised}   & 33.1       & 41.6 & 46.4 & 47.6 & 46.7 & 41.2 & 44.7 & 43.3 & 45.6 & 35.8 & 22.5 & \uline{40.8}  \\
		&           & STAGait\cite{ma2023fine} & 24.2       & 38.9 & 42.3 & 39.6 & 41.1 & 38.1 & 37.9 & 43.3 & 43.4 & 29.6 & 25.4 & 36.7          \\
		&           & Ours    & 28.7       & 49.8 & 55.7 & 57.1 & 54.7 & 51.2 & 49.8 & 48.2 & 48.4 & 37.7 & 28.5 & \textbf{46.4} \\
		GaitGL                                                                               & NM        & GOUDA\cite{habib2023watch}   & -          & -    & -    & -    & -    & -    & -    & -    & -    & -    & -    & \uline{92.8}  \\
		&           & UGRSF\cite{ren2023unsupervised}  & 84.2       & 95.7 & 96.2 & 94.7 & 89.9 & 87.1 & 89.1 & 91.8 & 96.7 & 93.2 & 63.6 & 89.3          \\
		&           & STAGait\cite{ma2023fine} & 76.7       & 88.3 & 88.8 & 87.1 & 88.4 & 82.9 & 85.9 & 90.6 & 93.1 & 93.9 & 73.0 & 86.2          \\
		&           & Ours    & 93.6       & 97.2 & 98.7 & 97.5 & 95.9 & 93.2 & 95.0 & 98.1 & 98.3 & 97.3 & 93.5 & \textbf{95.8} \\
		& BG        & GOUDA\cite{habib2023watch}   & -          & -    & -    & -    & -    & -    & -    & -    & -    & -    & -    & 80.9          \\
		&           & UGRSF\cite{ren2023unsupervised}  & 75.8       & 90.1 & 91.6 & 87.5 & 83.9 & 80.2 & 83.0 & 85.0 & 89.7 & 87.9 & 56.1 & \uline{82.8}  \\
		&           & STAGait\cite{ma2023fine} & 69.1       & 85.3 & 84.7 & 83.6 & 82.9 & 76.1 & 79.9 & 86.8 & 89.1 & 88.1 & 63.6 & 80.8          \\
		&           & Ours    & 88.9       & 95.9 & 96.9 & 93.1 & 89.8 & 86.2 & 88.9 & 94.6 & 96.3 & 96.0 & 88.8 & \textbf{92.3} \\
		& CL        & GOUDA\cite{habib2023watch}   & -          & -    & -    & -    & -    & -    & -    & -    & -    & -    & -    & 44.6          \\
		&           & UGRSF\cite{ren2023unsupervised}  & 53.6       & 71.1 & 76.9 & 75.5 & 72.9 & 69.1 & 69.8 & 69.1 & 71.7 & 60.7 & 31.2 & \uline{65.6}  \\
		&           & STAGait\cite{ma2023fine} & 43.9       & 65.3 & 66.5 & 66.9 & 68.8 & 60.0 & 66.4 & 71.2 & 69.3 & 61.8 & 42.2 & 62.0          \\
		&           & Ours    & 55.8       & 79.8 & 89.5 & 84.3 & 75.1 & 69.5 & 73.1 & 79.00 & 80.00 & 72.8 & 59.0 & \textbf{74.2} 
	\end{tblr}
\end{table}
\begin{table}[t]
\centering
\caption{Mean rank-1 (\%) comparison on CASIA-B. The best results of the model with different backbones in bold, and the second are marked with underline.}
\label{C-Mean}
\begin{tblr}{
  width = \linewidth,
  colspec = {Q[248]Q[496]Q[154]},
  cell{2}{1} = {r=4}{},
  cell{6}{1} = {r=4}{},
  hline{1-2,6,10} = {-}{},
}
Backbone & Method            & Mean          \\
GaitSet  & GOUDA\cite{habib2023watch}(WACV2024)   & 60.8          \\
         & UGRSF\cite{ren2023unsupervised}(Arxiv2023) & 71.3  \\
         & STANet\cite{ma2023fine}(ICCV2023)  & \uline{71.9} \\
         & \textbf{Ours}     & \textbf{72.3}         \\
GaitGL   & GOUDA\cite{habib2023watch}(WACV2024)   & 72.7          \\
         & UGRSF\cite{ren2023unsupervised}(Arxiv2023) & \uline{79.2}  \\
         & STANet\cite{ma2023fine}(ICCV2023)  & 76.3          \\
         & \textbf{Ours}     & \textbf{87.4} 
\end{tblr}
\end{table}

\definecolor{Alto}{rgb}{0.87,0.87,0.87}
\begin{table}
	\centering
	\caption{Rank-1 (\%) comparison on OUMVLP in various view angles. The best results are highlighted in bold, and the second are marked with underline.}
	\label{OUMVLP}
	\begin{tblr}{
			width = \linewidth,
			colspec = {Q[140]Q[180]Q[50]Q[50]Q[50]Q[50]Q[50]Q[50]Q[50]Q[52]Q[52]Q[52]Q[52]Q[52]Q[52]Q[52]Q[58]},
			row{3} = {Alto,c},
			row{7} = {Alto,c},
			row{12} = {Alto,c},
			cell{1}{1} = {r=2}{},
			cell{1}{2} = {r=2}{},
			cell{1}{3} = {c=14}{0.714\linewidth,c},
			cell{1}{17} = {r=2}{},
			cell{3}{1} = {c=17}{0.92\linewidth},
			cell{7}{1} = {c=17}{0.92\linewidth},
			cell{8}{1} = {r=2}{},
			cell{10}{1} = {r=2}{},
			cell{12}{1} = {c=17}{0.92\linewidth},
			cell{13}{1} = {r=2}{},
			cell{15}{1} = {r=2}{},
			hline{1,3-4,7-8,10,12-13,15,17} = {-}{},
			hline{2} = {3-16}{},
		}
		Backbone                                                                            & Method  & Probe View &      &      &      &      &      &      &      &      &      &      &      &      &      & Mean          \\
		&         & 0°         & 15°  & 30°  & 45°  & 60°  & 75°  & 90°  & 180° & 195° & 210° & 225° & 240° & 255° & 270° &               \\
		Fully supervised methods with all labels on OUMVLP                                  &         &            &      &      &      &      &      &      &      &      &      &      &      &      &      &               \\
		& GaitNet\cite{zhang2019gait} & 11.4       & 29.1 & 41.5 & 45.5 & 39.5 & 41.8 & 38.9 & 14.9 & 33.1 & 43.2 & 45.6 & 39.4 & 40.5 & 36.3 & 35.8          \\
		& GaitSet\cite{chao2021gaitset} & 79.5       & 87.9 & 89.9 & 90.2 & 88.1 & 88.7 & 87.8 & 81.7 & 86.7 & 89.0 & 89.3 & 87.2 & 87.8 & 86.2 & 87.1          \\
		& GaitGL\cite{lin2022gaitgl}  & 91.7       & 92.6 & 92.3 & 92.5 & 95.8 & 92.2 & 92.1 & 92.4 & 91.9 & 91.7 & 91.9 & 92.1 & 91.5 & 91.4 & 92.0          \\
		Unsupervised Domain Adaptation methods without any labels on OUMVLP (CASIA-B $\rightarrow$ OUMVLP) &         &            &      &      &      &      &      &      &      &      &      &      &      &      &      &               \\
		GaitSet                                                                             & GOUDA\cite{habib2023watch}   & -          & -    & -    & -    & -    & -    & -    & -    & -    & -    & -    & -    & -    & -    & \uline{27.9}          \\
		& Ours    & 24.6       & 44.1 & 50.9 & 51.1 & 45.5 & 46.1 & 43.2 & 28.2 & 44.4 & 49.8 & 50.8 & 44.6 & 44.7 & 39.3 & \textbf{43.6} \\
		GaitGL                                                                              & GOUDA\cite{habib2023watch}   & -          & -    & -    & -    & -    & -    & -    & -    & -    & -    & -    & -    & -    & -    & \uline{34.1}          \\
		& Ours    & 28.9       & 48.3 & 55.4 & 55.9 & 51.4 & 52.3 & 48.8 & 30.5 & 44.9 & 53.1 & 53.1 & 48.9 & 47.6 & 43.8 & \textbf{49.2} \\
		Unsupervised Domain Adaptation methods without any labels on OUMVLP (GREW $\rightarrow$  OUMVLP)    &         &            &      &      &      &      &      &      &      &      &      &      &      &      &      &               \\
		GaitSet                                                                             & GOUDA\cite{habib2023watch}   & -          & -    & -    & -    & -    & -    & -    & -    & -    & -    & -    & -    & -    & -    & \uline{38.8}          \\
		& Ours    & 47.9       & 68.3 & 76.2 & 78.5 & 69.5 & 68.2 & 64.2 & 51.6 & 68.7 & 75.7 & 78.8 & 73.4 & 71.7 & 66.3 & \textbf{68.5} \\
		GaitGL                                                                              & GOUDA\cite{habib2023watch}   & -          & -    & -    & -    & -    & -    & -    & -    & -    & -    & -    & -    & -    & -    & \uline{44.2}          \\
		& Ours    & 40.5       & 58.0 & 65.4 & 67.3 & 63.1 & 64.1 & 60.8 & 44.1 & 58.4 & 67.4 & 67.9 & 65.8 & 63.9 & 61.7 & \textbf{60.6} 
	\end{tblr}
\end{table}

\subsection{Comparison with SOTA Methods}
\subsubsection{Evaluation on CASIA-B}
In this experiment, we use OUMVLP which contains a large number of subjects and different viewing angles, as the source dataset to pre-train the feature extractor. The performance of different clothing on the CASIA-B dataset is shown in Table \ref{CAISA-B}, and the mean is shown in Table \ref{C-Mean} . Our method achieves superior performance when using either GaitSet or GaitGL as the backbone network. Specifically, our model improves accuracy by 5.6\% in CL with backbone GaitSet, demonstrating its effectiveness in complex dressing scenarios. Notably, when using GaitGL as the backbone, our method achieves excellent performance in all walking conditions, with improvements of 3.0\% and 8.9\% in NM and BG, respectively. In the most challenging CL scenario, the accuracy improved by 8.6\%, reaching 74.2\%. In addition, the average accuracy of our method in the three dressing situations reached 87.4\%, which is 8.2\% higher than the SOTA method, and even surpassing some fully supervised learning methods.

The exceptional performance of our approach can benefit from a detailed analysis of the specific challenges in unsupervised gait recognition compared to previous unsupervised gait recognition. Two primary challenges were identified: suboptimal clustering results and the presence of noisy pseudo-labels. To address these issues, we introduced dynamic clustering techniques to enhance clustering accuracy and employed contrastive learning to refine pseudo-labels. The experimental results strongly validate the effectiveness of our proposed solutions, demonstrating the model's capability to overcome these challenges and achieve superior performance in unsupervised gait recognition tasks.

Notably, our method achieves particularly significant improvements under the challenging "CL" condition, where clothing variations present substantial intra-class differences. This highlights the critical role of pseudo-label refinement in addressing label noise and improving model accuracy. Furthermore, the experimental results reveal that using the GaitGL backbone consistently yields better accuracy compared to GaitSet in all tested conditions. This suggests that the choice of backbone architecture plays a pivotal role in feature extraction quality and directly impacts the model’s overall performance. The consistent improvements observed with GaitGL emphasize the importance of leveraging more advanced and capable backbones in unsupervised gait recognition frameworks.

\subsubsection{Evaluation on OUMVLP}
In order to verify the effect of our proposed method, we conduct a series of unsupervised and adaptive experiments on the OUMVLP dateset using CASIA-B and GREW datasets as source datasets, respectively, and the results are shown in Table \ref{OUMVLP}. By comparing the results obtained with different source datasets—CASIA-B and GREW—we observe notable variations that highlight the importance of source dataset quality and diversity. When using CASIA-B as the source dataset, GaitDCCR achieves a mean rank-1 accuracy of 43.6\% with the GaitSet backbone and 49.2\% with the GaitGL backbone. These results already surpass many existing unsupervised domain adaptation methods. However, when the GREW dataset is used as the source, the performance improves significantly, with a mean rank-1 accuracy reaching 68.5\% with GaitSet and 60.6\% with GaitGL. This stark difference underscores the critical role of dataset size and variability: GREW, with its larger subject pool and richer semantic information, enables better feature extraction and domain adaptation compared to CASIA-B, which has relatively limited diversity. In this experiment, we did not conduct comparative experiments with UGRSF and STAGait. The reason is that UGRSF has not been tested on the OUMVLP dataset and its code is not publicly available. Similarly, STAGait only utilizes a subset of OUMVLP in its experiments, which is inconsistent with the full-dataset evaluation performed in our study.

The results also highlight the advantages of our method over GOUDA, a state-of-the-art approach. For instance, with the GREW dataset as the source, the proposed GaitDCCR achieves a mean rank-1 accuracy improvement of over 30\% compared to GOUDA, demonstrating the effectiveness of our dynamic clustering and label refinement strategies. Similarly, when using CASIA-B as the source, proposed GaitDCCR still outperforms GOUDA, further validating the robustness of our approach even when trained on a less diverse dataset. These findings firmly establish proposed GaitDCCR as a robust and effective solution for unsupervised gait recognition in large-scale datasets like OUMVLP. The results also emphasize the importance of leveraging diverse and comprehensive source datasets to achieve optimal domain adaptation performance.

\subsection{Ablation Studies}
Following the experimental setup of previous methods, the ablation studies of the model were done on the public CASIA-B dataset.
\subsubsection{Effectiveness of DCP and DWC in the clustering phase}
\begin{figure}[t]
	\centering
	\includegraphics[width=\linewidth]{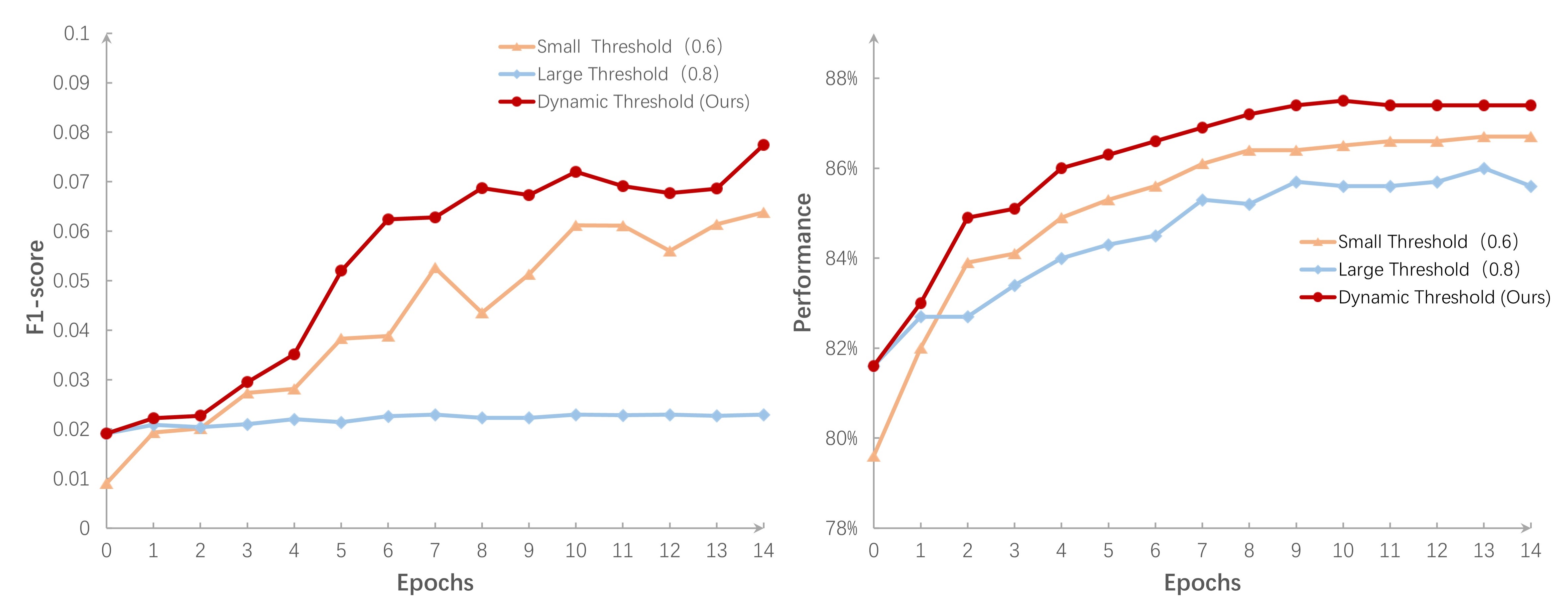}
	\caption{Comparison of previous fixed small and large  probability threshold and our dynamic threshold in terms of F1-score and performance.}
	\label{ACC}
\end{figure}

To evaluate the effectiveness of the proposed methods in the clustering stage, we performed ablation studies by incrementally adding the Dynamic Clustering Parameters (DCP) and Dynamic Weighted Centroids (DWC) to the baseline. The results, summarized in Table \ref{Commponent}, demonstrate that both DCP and DWC contribute to notable performance improvements. To further substantiate the impact of these two components, we conducted quantitative analyses. Firstly, in order to evaluate the effectiveness of the proposed dynamic clustering parameters (DCP) in reducing the impact of noisy samples, we compare it with the fixed smaller threshold $\epsilon$ (0.6) and larger $\epsilon$ (0.8) methods in terms of $F1$ scores and overall performance. As shown in the Figure \ref{ACC}, the proposed dynamic threshold method shows higher $F1$ scores than the fixed small $\epsilon$ method at the early stage of training, and continues to improve and eventually reaches the highest level as the training progresses. At the same time, the model performance is significantly better than the other two methods. In contrast, the fixed small $\epsilon$ fails to effectively filter the noisy samples at the early stage of training, resulting in the $F1$ scores and performance not reaching the desired level at the later stage, while the fixed large $\epsilon$ limits the $F1$ scores and performance improvement of the model due to the over- exclusion of real samples in the clusters at the later stage of training. This demonstrates that the proposed DCP method effectively reduces the number of noisy samples in clustering, optimizes clustering results, and improves overall model performance.

\begin{figure}[t]
	\centering
	\includegraphics[width=\linewidth]{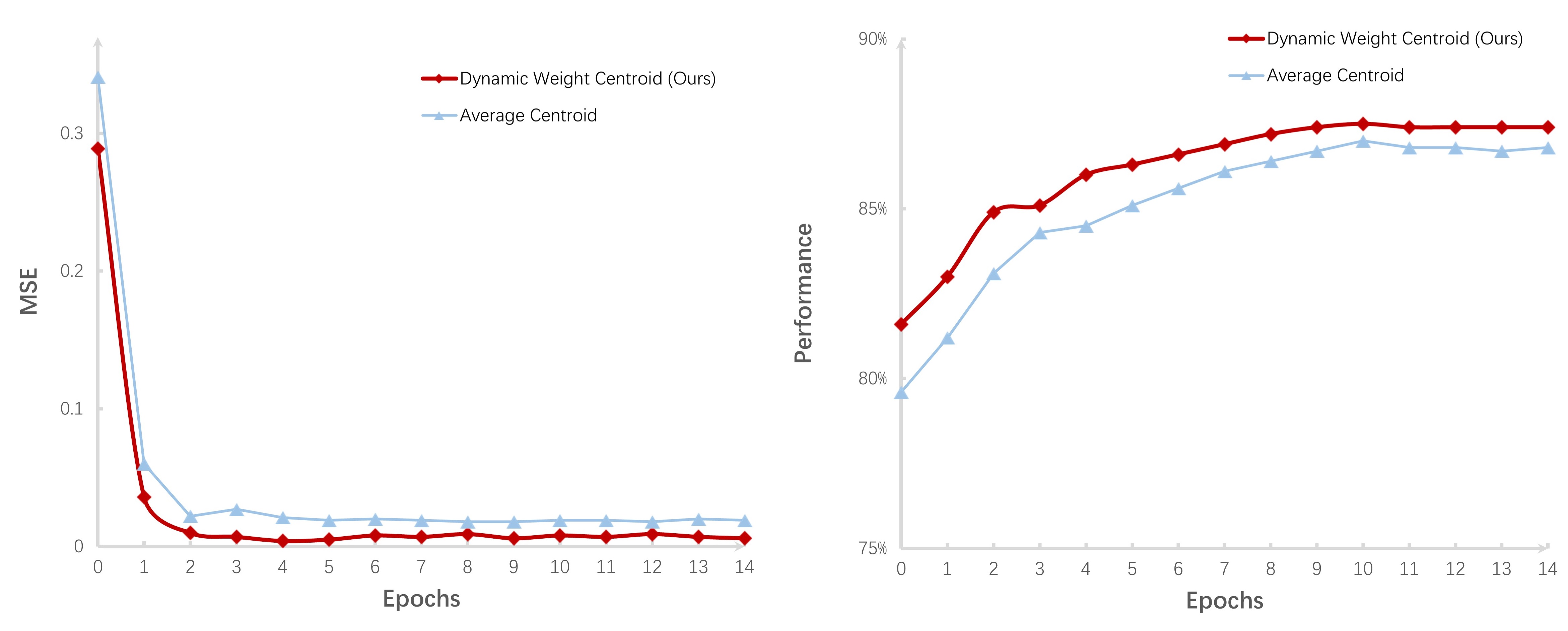}
	\caption{Comparison of the previous average centroid and our dynamic weighted centroid in terms of mean square error and performance.}
	\label{MSE}
\end{figure}

Secondly, to compare the accuracy of the cluster centroids computed using DWC with the average centroids used in previous methods, we measured the mean square error (MSE) between the computed centroids and the true centroids of the clusters. The true centroids were defined as the mean feature vectors of correctly clustered samples. As illustrated in Figure \ref{MSE}, DWC produces more accurate centroids, particularly during the early stages of training when the model's feature extraction ability is still developing. This improvement highlights the ability of DWC to effectively mitigate the influence of noisy samples by assigning dynamic weights based on sample density within clusters.

Overall, these findings underscore the importance of DCP and DWC in enhancing clustering performance. By dynamically adapting clustering parameters and refining centroid calculations, these components significantly improve the reliability of the clustering process, laying a strong foundation for subsequent training phases.
\definecolor{Mercury}{rgb}{0.909,0.909,0.909}
\begin{table}[t]
\centering
\caption{Effectiveness of each component in our framework. "DCP" stands for dynamic clustering parameters,“DWC” represents dynamic weighted centroids, “CPR” stands for confidence-based pseudo-label refinement, "DA" Indicates data augmentation, and “CTM” stands for contrastive teacher module with "DA". }
\label{Commponent}
\begin{tblr}{
		width = \linewidth,
		colspec = {Q[121]Q[69]Q[81]Q[65]Q[75]Q[202]Q[67]Q[67]Q[67]Q[88]},
		cells = {c},
		row{4} = {Mercury},
		row{8} = {Mercury},
		row{12} = {Mercury},
		cell{1}{1} = {r=2}{},
		cell{1}{2} = {c=5}{0.492\linewidth},
		cell{1}{7} = {r=2}{},
		cell{1}{8} = {r=2}{},
		cell{1}{9} = {r=2}{},
		cell{1}{10} = {r=2}{},
		cell{3}{2} = {c=5}{0.492\linewidth},
		cell{4}{1} = {c=10}{0.901\linewidth},
		cell{8}{1} = {c=10}{0.901\linewidth},
		cell{12}{1} = {c=10}{0.901\linewidth},
		hline{1,3-15} = {-}{},
		hline{2} = {2-6}{},
		}
		Index                              & Compoent &     &     &     &               & NM   & BG   & CL   & Mean \\
		& DCP      & DWC & CPR & CTM & CTM  (w/o DA) &      &      &      &      \\
		Baseline                           &          &     &     &     &               & 94.2 & 88.5 & 43.9 & 75.5 \\
		Components of the Clustering Phase &          &     &     &     &               &      &      &      &      \\
		1                                  & \checkmark        &     &     &     &               & 95.4 & 90.7 & 46.3 & 77.5 \\
		2                                  &          & \checkmark   &     &     &               & 96.2 & 90.2 & 46.7  & 77.6 \\
		3                                  & \checkmark        & \checkmark   &     &     &               & 96.1 & 90.5 & 49.9 & 78.8 \\
		Components for the Training Phase  &          &     &     &     &               &      &      &      &      \\
		4                                  &          &     & \checkmark   &     &               & 96.5 & 92.3 & 54.9 & 81.5 \\
		5                                  &          &     &     & \checkmark   &               & 96.1 & 91.6 & 60.2 & 82.6 \\
		6                                  &          &     & \checkmark   & \checkmark   &               & 95.1 & 91.8 & 67.9 & 85.6 \\
		All Components                     &          &     &     &     &               &      &      &      &      \\
		7                                  & \checkmark        & \checkmark   & \checkmark   &     & \checkmark             & 96.8 & 92.5 & 65.9 & 85.1 \\
		8                                  & \checkmark        & \checkmark   & \checkmark   & \checkmark   &               & 95.8 & 92.3 & 74.2 & 87.4 
	\end{tblr}
\end{table}

\subsubsection{Effectiveness of CPR and CTM in the clustering phase}
To validate the effectiveness of the proposed components in training phase, we sequentially add Confidence-Based Pseudo Label Refinement (CPR), and Contrastive Teacher Module (CTM) to the baseline, as shown in Table \ref{Commponent}. Experimental results show that the components proposed in the clustering stage can greatly improve the accuracy of the model, and the improvement effect is more significant than the components proposed in the clustering stage. The above results illustrate that it is crucial to refine the noisy pseudo-labels in the clustering stage in clustering-based unsupervised gait recognition, and our model can effectively solve this problem.

Secondly, to evaluate the effectiveness of our proposed data augmentation method, we conducted an experiment where the input to the Contrastive Teacher Module (CTM) was provided without data augmentation. The results, as shown in Table \ref{Commponent} (experiments "7" and "8"), indicate that the proposed data augmentation strategy significantly enhances model performance, particularly in addressing challenges related to clothing variation. By simulating different clothing scenarios, the augmentation method allows the model to capture more diverse features, improving its robustness and generalization.

\begin{table}[t]
\centering
\caption{Mean rank-1 (\%) comparison of different clustering parameter decay methods on CASIA-B. The best results in bold.}
\label{eps}
\begin{tblr}{
  width = \linewidth,
  colspec = {Q[246]Q[150]Q[150]Q[150]Q[183]},
  hline{1-2,5} = {-}{},
}
Method       & NM            & BG            & CL            & Mean          \\
Square         & 95.1          & 91.6          & 68.1          & 84.9          \\
Linear       & 96.2          & 92.0          & 68.6          & 85.6         \\
\textbf{Exponential} & \textbf{95.8} & \textbf{92.3} & \textbf{74.2} & \textbf{87.4} 
\end{tblr}
\end{table}

\begin{figure}[t]
	\centering
	\includegraphics[width=\linewidth]{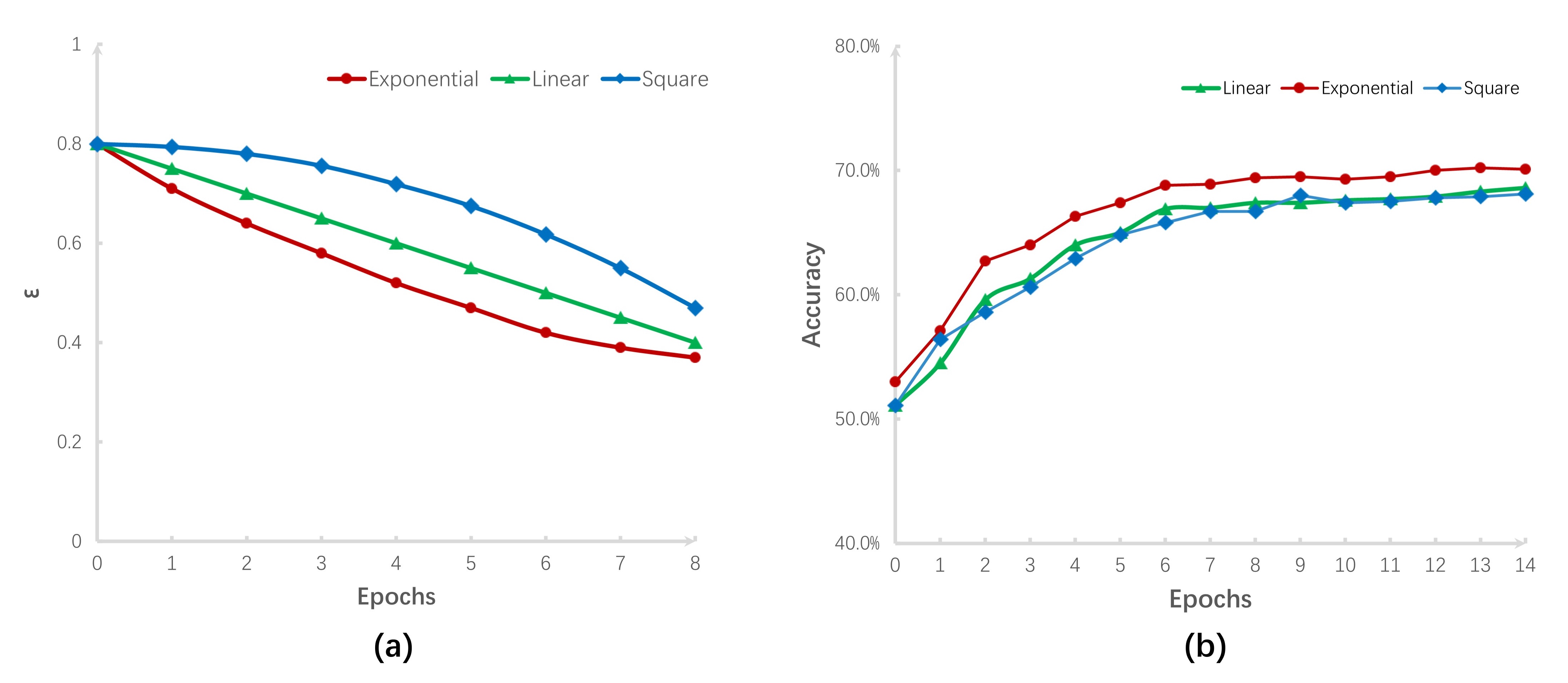}
	\caption{(a) Different clustering parameter decay methods in DCP. (b) The accuracy of CL clothing  during training.}
	\label{cl}
\end{figure}

\subsubsection{Effectiveness of different dynamic clustering parameters}
In this section, we evaluate the impact of different parameter decay methods in DCP
on model performance. In the experiment, square decay, linear decay and exponential decay,  the experimental results are shown in the Table \ref{eps}. From this result, the exponential decay method achieves the best effect, and the square decay method gives the worst effect. We believe that this is related to the learning ability of the model during the training process. Figure \ref{cl} shows the change in $\epsilon$ and CL accuracy for different parameter decay methods. It can be seen that CL accuracy grow faster in the early stage of the  training process and then gradually grow slowly, and the exponential decay method is similar to the above change, while the square decay method is just the opposite.

\begin{table}[t]
\centering
\caption{The Impact of the number of cluster ($k$-value) in the Latent Cluster Set on the Model.}
\label{knn}
\begin{tblr}{
  width = \linewidth,
  colspec = {Q[288]Q[185]Q[185]Q[185]},
  hline{1-2,6} = {-}{},
}
Setting & NM   & BG   & CL   \\
k=1     & 90.5 & 86.4 & 67.8 \\
\textbf{k=2}     &\textbf{95.8} & \textbf{92.3} & \textbf{74.2} \\
k=3     & 95.2 & 91.4 & 70.4 \\
k=4     & 94.8 & 89.9 & 68.1 
\end{tblr}
\end{table}
\subsubsection{Impact of $k$ in Latent Cluster Set}
In this experiment, we explore how the number of cluster ($k$-value) in the teacher branch's Latent Cluster Set specifically impact model performance. Results shown in Table \ref{knn} reveals that setting $k=2$ leads to the best  performance. This finding reveals the significant impact of the data augmentation strategy's effectiveness on the teacher branch's performance. If $k$ is set too low, it might be challenging to closely match clusters formed due to different clothing; conversely, if $k$ is set too high, it could result in proximity to unrelated clusters. Therefore, $ k=2 $ is set as the optimal choice in our experiments. This insight is vital for optimizing designs of models based on the KNN algorithm, emphasizing the importance of precise $k$ value selection during data augmentation and model training processes.

\section{Conclusion}
This paper identifies a prevalent challenge in unsupervised gait recognition: inaccurate clustering. This issue arises due to a number of complex factors, including clothing and angle changes, which cause samples with different identities are grouped together and samples with the same identity are separated into different clusters. This results in the generation of noisy pseudo-labels. Furthermore, we optimized two major issues with existing unsupervised gait recognition methods that directly use noisy pseudo-labels for training: unreliable clustering centroids and one-hot label strategy.

To address these challenges, we propose a new model called GaitDCCR. Firstly, in the clustering stage, we improve the efficiency of clustering by dynamically adjusting the clustering parameters, and recalculate the clustering center by assigning weights to each sample. Secondly, in the training stage, we adopt the classic Teacher-student structure to reduce the impact of noisy pseudo-labels. Specifically, in the student branch, a confidence matrix is calculated to store the confidence that each sample belongs to all clusters, which is not used to obtain confidence-based pseudo-labels. In the teacher branch, each sample is augmented to simulate variations in clothing conditions, and then finds clusters with different clothing from the original sample in the feature space. Extensive experiments have demonstrated that our model has achieved excellent results in unsupervised domain adaptation experiments on CASIA-B and OUMVLP datasets. Specifically, when using GaitGL as the backbone on  CASIA-B, our model outperforms all SOTA methods and even some fully supervised methods. The results proves that the proposed method is simple and effective, it can effectively reduce the impact of noisy pseudo-labels. Furthermore, the results of CASIA-B to OUMVLP reveals deficiencies in unsupervised domain adaptation techniques. These methods require a significant amount of semantic knowledge from the source dataset to ensure that the pre-trained model acquires a substantial amount of valuable information.

In the future, our objective is to develop more efficient and accurate gait recognition technologies that can handle a wide range of complex factors. It will further enhance the resilience and adaptability of gait recognition in real-world.

\section{Acknowledgments}
This work is funded by National Natural Science Foundation of China (62002215) and Shanghai Pujiang Program (20PJ1404400)

\bibliographystyle{elsarticle-num}
\bibliography{ref}

\end{document}